\documentclass{article}

\usepackage[preprint, nonatbib]{neurips_2026}

\usepackage[utf8]{inputenc}
\usepackage[T1]{fontenc}
\usepackage{hyperref}
\usepackage{url}
\usepackage{booktabs}
\usepackage{amsfonts}
\usepackage{nicefrac}
\usepackage{microtype}
\usepackage{xcolor}

\usepackage{graphicx}
\usepackage{float}
\usepackage{amsmath, amssymb}
\usepackage{tikz}
\usetikzlibrary{shapes.geometric, arrows, positioning}

\title{RLearner-LLM: Balancing Logical Grounding and Fluency in Large Language Models via Hybrid Direct Preference Optimization}

\author{%
  Qiming Bao \\
  University of Auckland \\
  \texttt{qbao775@aucklanduni.ac.nz}
  \And
  Juho Leinonen \\
  Aalto University \\
  \texttt{juho.2.leinonen@aalto.fi}
  \And
  Paul Denny \\
  University of Auckland \\
  \texttt{p.denny@auckland.ac.nz}
  \And
  Michael J.\ Witbrock \\
  University of Auckland \\
  \texttt{m.witbrock@auckland.ac.nz}
}

\begin{document}
\maketitle

\begin{abstract}
Direct Preference Optimization (DPO) has emerged as an efficient alternative to PPO-based RLHF, yet its application to knowledge-intensive generation tasks reveals a fundamental limitation: standard preference signals---whether from human annotators or LLM judges---exhibit a systematic \textit{verbosity bias} that rewards linguistic fluency over logical correctness. We show empirically that this reward signal blindspot leaves a large logical alignment gap: SFT-trained models achieve NLI entailment scores of only 0.05--0.22, despite producing fluent, confident-sounding text. We propose \textbf{RLearner-LLM}, a framework that resolves this through \textbf{Hybrid-DPO}: an automated preference construction pipeline that fuses a DeBERTa-v3 NLI entailment signal with a verifier LLM score, eliminating the need for human annotation while overcoming the ``alignment tax'' that degrades fluency under single-signal optimization. Evaluated across five academic domains (Biology, Medicine, Law) with three base architectures (LLaMA-2-13B, Qwen3-8B, and Gemma 4 E4B-it), RLearner-LLM achieves up to 6$\times$ NLI improvement over SFT baselines, with NLI gains in $11$ of $15$ (architecture, domain) cells and consistent answer-coverage gains. On Gemma 4 E4B-it---a 4.5B-effective-parameter dense model---Hybrid-DPO lifts NLI entailment in four of five domains (from +11.9\% on Auckland Law to +2.4$\times$ on UK Medicine Year 2), with faster inference across all five, demonstrating that the approach scales down to compact modern base models without losing the alignment-tax mitigation. We further show these gains are robust to aggressive, response-grounded question-quality filtering: under a \textit{strict-discriminator} filter that keeps only questions whose author key matches the modal student answer (computed from per-option student response logs), answer-coverage improves in every architecture--corpus cell and NLI in four of six. Our Qwen3-8B RLearner-LLM wins $95\%$ of blind pairwise comparisons against its own SFT baseline, while GPT-4o-mini in turn wins $95\%$ against our concise RLearner-LLM output---a result that, viewed alongside the $69\%$ win rate the same judge awards a verbose SFT baseline over our DPO-aligned model, reproduces the verbosity bias we document on a frontier comparator and reinforces the case for logic-aware automatic metrics (NLI, ACR) over LLM-as-a-judge on knowledge-intensive generation.\footnote{Code, training scripts, and per-domain evaluation artefacts: \url{https://github.com/14H034160212/Explanation-Generation}}
\end{abstract}

\section{Introduction}

The generation of high-quality educational explanations represents a fundamental challenge in natural language processing. Unlike general-purpose conversational tasks, educational rationales must simultaneously satisfy two competing requirements: they must be \textit{logically grounded}---deriving the correct answer through sound deductive steps---and \textit{linguistically fluent}---maintaining natural, readable prose appropriate for a learner audience.

Supervised Fine-Tuning (SFT) on teacher-generated demonstrations has become the standard approach to instilling this capability into LLMs. However, SFT models exhibit a consistent and critical failure mode: while they produce fluent, confident-sounding text, they rarely construct the deductive chain that justifies \textit{why} the correct answer follows from the question context. Empirically, SFT models in our study achieve NLI entailment scores of only 0.05--0.22, indicating that the generated explanation rarely logically implies the ground-truth answer---even when the model ``knows'' the fact.

Reinforcement Learning from Human Feedback (RLHF) and its efficient derivative, Direct Preference Optimization (DPO) \cite{rafailov2023direct}, have sought to move beyond SFT by aligning models to human preferences. Yet these approaches contain a structural flaw when applied to factually demanding tasks: \textbf{the preference signal does not measure logical correctness}. Human annotators---and LLM judges---exhibit a well-documented \textit{verbosity bias}, systematically preferring longer, rhetorically polished responses over concise, logically precise ones \cite{zheng2024judging}. As we demonstrate empirically, GPT-4o-mini awards a 69\% win rate to a verbose SFT baseline over a more logically precise but concise RL-aligned model. DPO trained on such signals therefore reinforces stylistic fluency at the expense of factual grounding, leaving the core alignment gap unaddressed.

A natural remedy is to replace human preference labels with an automated logical metric---specifically, NLI entailment probability. However, optimizing DPO solely against NLI entailment incurs what we term the \textbf{alignment tax}: the model learns to produce short, repetitive snippets that satisfy the NLI scorer but collapse in linguistic coherence. This trade-off between logical soundness and readability represents a fundamental tension that single-dimensional reward signals cannot resolve.

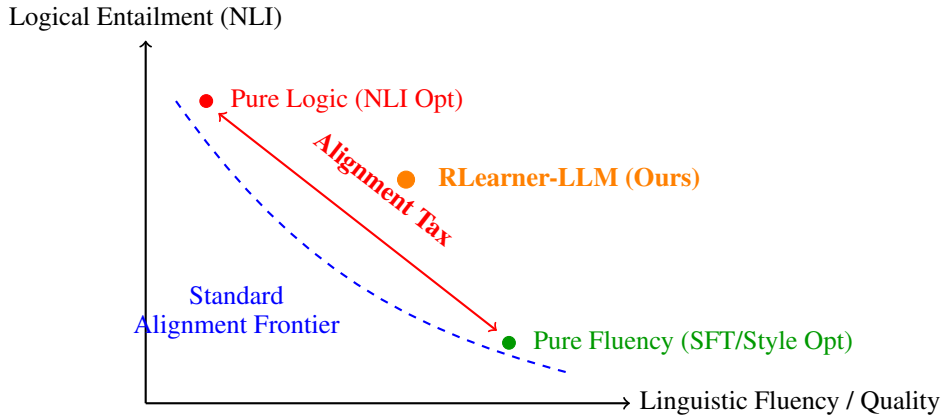
\begin{figure}[h]
\centering
\begin{tikzpicture}[scale=0.8]
\draw[->, thick] (0,0) -- (8,0) node[right] {Linguistic Fluency / Quality};
\draw[->, thick] (0,0) -- (0,6) node[above] {Logical Entailment (NLI)};

\draw[blue, thick, dashed] (0.5,5) to[bend right=20] (7,0.5);
\node[blue, align=center] at (1.5,1.5) {Standard\\Alignment Frontier};

\filldraw[red] (1,5) circle (3pt) node[right=0.2cm] {Pure Logic (NLI Opt)};
\filldraw[green!60!black] (6,1) circle (3pt) node[right=0.2cm] {Pure Fluency (SFT/Style Opt)};
\filldraw[orange] (4.3,3.7) circle (4pt) node[right=0.3cm] {\textbf{RLearner-LLM (Ours)}};

\draw[<->, red, thick] (1.2,4.8) -- (5.8,1.2) node[pos=0.5, sloped, above=0.25cm] {\textbf{Alignment Tax}};

\end{tikzpicture}
\caption{Conceptual illustration of the \textbf{Alignment Tax}. Standard optimization forces a trade-off between logical grounding and linguistic fluency. RLearner-LLM pushes the Pareto frontier toward the upper-right quadrant through a dual-signal hybrid reward.}
\label{fig:alignment_tax}
\end{figure}

In this paper, we propose \textbf{RLearner-LLM}, a framework that resolves this tension by addressing the root cause directly: the reward signal used to construct preference pairs. Our key insight is that logical grounding and linguistic fluency are \textit{complementary objectives}, not competing ones---and that an automated dual-signal reward combining NLI entailment (for logical soundness) with a verifier score (for pedagogical quality) can construct high-fidelity preference pairs without human annotation, while simultaneously avoiding the degenerate solutions that plague single-signal optimization.

Our contributions are as follows:
\begin{enumerate}
    \item \textbf{Problem diagnosis}: We identify that the root cause of logical misalignment in DPO-trained educational models is not the optimization algorithm itself, but the \textit{reward signal blindspot}---the inability of standard preference signals to measure logical correctness. We provide empirical evidence of verbosity bias in LLM-as-a-judge evaluation that perpetuates this blindspot.
    \item \textbf{Hybrid-DPO framework}: We introduce \textbf{RLearner-LLM}, which automatically constructs preference pairs $\mathcal{D}_{pref}$ using a composite score $H(E_i) = 0.5\cdot S_{\text{NLI}}(E_i) + 0.5\cdot \tilde{S}_{\text{Verifier}}(E_i)$, fusing a DeBERTa-v3 NLI signal with a min-max-normalised verifier LLM score (Eq.~\ref{eq:hybrid_reward}). The dual-signal hybrid eliminates the need for human annotation while overcoming the alignment tax.
    \item \textbf{Cross-architecture validation}: We validate across three base architectures (LLaMA-2-13B, Qwen3-8B, and Gemma 4 E4B-it) and five academic domains (Biology, Medicine, Law), demonstrating up to 6$\times$ NLI improvement over SFT baselines in $11$ of $15$ (architecture, domain) cells, with consistent ACR gains on stronger base models. Gemma 4 E4B-it---a 4.5B-effective-parameter dense model---shows Hybrid-DPO NLI gains in four of five domains (+11.9\% Auckland Law, +32.0\% UK Medicine Y1, +65.6\% Cardiff Biology, +2.4$\times$ UK Medicine Y2), with faster inference across all five and the first single-pass-RL result to surpass the iterative ILearner-LLM (K=5) baseline on Auckland Law NLI. Our Qwen3-8B model further wins $95\%$ of blind pairwise comparisons against its own SFT baseline; on the same judge it loses $95\%$ to GPT-4o-mini's longer outputs, replicating the verbosity bias on a frontier comparator and motivating logic-aware automatic metrics over LLM-as-a-judge for knowledge-intensive generation.
    \item \textbf{Robustness to question-quality filtering}: Using the full PeerWise answer-submission logs, we show the gains survive a \textit{strict-discriminator} filter that keeps only questions whose author key matches the \emph{modal student answer}: across three architectures and two corpora, answer-coverage improves in all six (architecture, corpus) cells and NLI in four of six, reproducing the per-architecture pattern of the main results (Appendix~\ref{sec:tierc_appendix}).
\end{enumerate}

\section{Related Work}
\label{sec:related_work}

\subsection{Alignment and Preference Optimization}
Optimizing Large Language Models (LLMs) to align with human intent has predominantly relied on Reinforcement Learning from Human Feedback (RLHF) \cite{ouyang2022training}, utilizing Proximal Policy Optimization (PPO) \cite{schulman2017proximal}. Recently, Direct Preference Optimization (DPO) \cite{rafailov2023direct} emerged as a stable alternative that mathematically bypasses the need for an explicit reward model. Building on this, recent advancements like RRHF \cite{yuan2023rrhf}, ORPO \cite{hong2024orpo}, and SimPO \cite{meng2024simpo} have further simplified alignment via reference-free objectives. The state-of-the-art in 2025 has moved toward adaptive batch-wise scheduling \cite{zhang2025adaptive} and iterative dual-based methods for constrained alignment \cite{chen2025constrained}, addressing the inherent algorithmic biases identified in standard RLHF \cite{ji2024limits, shrivastava2024algorithmic}.

\subsection{Hallucination Mitigation and Reasoning}
In knowledge-intensive tasks, such as educational explanation generation, factual correctness is paramount. Prior work has explored grounding LLMs using retrieval-augmented generation (RAG) \cite{lewis2020retrieval} or process-based feedback \cite{uesato2023verify}. Modern techniques focus on mechanistic hallucination detection \cite{li2023haludetect} and Multi-Model Contrastive Decoding (MCD) \cite{liu2025contrastive}. While debates continue regarding whether emergent abilities are a metric-driven "mirage" \cite{schaeffer2023mirage}, advancements in 2025 demonstrate that reasoning quality can be significantly enhanced by verifying "decision pivots" in Chain-of-Thought traces \cite{wang2025pivots} or equipping models with cognitive tools \cite{zhao2025cognitive}. However, these logical optimizations often degrade fluency—the "alignment tax"—which our work addresses via a hybrid verifier-consistent reward.

\subsection{LLM-as-a-Judge and Evaluation Bias}
The use of powerful instruction-tuned models like GPT-4 as automated evaluators \cite{zheng2024judging} is widespread but prone to significant biases, including verbosity bias \cite{wang2023large} and self-preference bias \cite{pan2024evalbias}. The verbosity-bias confound has been formally analysed in DPO itself: \cite{park2024lengthbias} show that standard DPO and its $\beta$-tuned variants \cite{park2024betadpo} systematically conflate explanation length with quality, and propose length-aware regularisation as a partial remedy. Our work targets the \emph{upstream} reward-construction stage rather than the downstream loss: by replacing length-correlated stylistic preferences with an NLI-anchored dual-signal reward (Section~\ref{sec:first_principles}) we cut the bias at its origin, which the head-to-head Table~\ref{tab:pairwise_1} confirms persists on frontier comparators despite length controls applied at the output stage. Comprehensive assessments like DecodingTrust \cite{wang2023trust} further highlight the vulnerability of current models to stylistic manipulation; we maintain safety and knowledge-retention standards \cite{kim2025safety} throughout.

\definecolor{NBlue}{HTML}{1A56A0}
\definecolor{NRed}{HTML}{C0392B}
\definecolor{NGreen}{HTML}{1A7A45}
\definecolor{NGold}{HTML}{B07D1A}
\definecolor{NSlate}{HTML}{4A5568}
\definecolor{NBlueFill}{HTML}{EBF2FB}
\definecolor{NRedFill}{HTML}{FBEAEA}
\definecolor{NGreenFill}{HTML}{EBF5EF}
\definecolor{NGoldFill}{HTML}{FDF6E3}
\definecolor{NSlateFill}{HTML}{F0F2F5}
\definecolor{NStroke}{HTML}{BBBDC2}

\begin{figure}[t!]
\centering
\begin{tikzpicture}[
    scale=0.88,
    >=stealth,
    iobox/.style={
        draw=NStroke, fill=NSlateFill,
        rounded corners=4pt, align=center,
        font=\small, inner sep=7pt, line width=0.6pt
    },
    modbox/.style={
        draw=#1, fill=#1Fill,
        rounded corners=5pt, align=center,
        font=\small\bfseries, inner sep=7pt,
        minimum height=1.0cm, minimum width=3.2cm,
        line width=0.7pt
    },
    rewardbox/.style={
        draw=NGold, fill=NGoldFill,
        rounded corners=5pt, align=center,
        font=\footnotesize, inner sep=10pt,
        minimum height=1.7cm, minimum width=11.0cm,
        line width=0.7pt
    },
    dpobox/.style={
        draw=NBlue, fill=NBlueFill,
        rounded corners=5pt, align=center,
        font=\small\bfseries, inner sep=8pt,
        minimum height=1.0cm, minimum width=9.0cm,
        line width=0.8pt
    },
    arr/.style={->, line width=0.7pt, color=NSlate, shorten >=3pt, shorten <=3pt},
    arrd/.style={->, line width=0.7pt, color=NBlue!50, dashed, shorten >=3pt, shorten <=3pt},
    steplabel/.style={font=\footnotesize\bfseries, color=NSlate},
    sublabel/.style={font=\scriptsize, color=NSlate!80}
]

\draw[draw=NStroke!50, fill=NSlateFill!40, rounded corners=8pt, line width=0.5pt, dashed]
    (-5.6, 2.8) rectangle (5.6, 5.0);
\node[sublabel, anchor=north west] at (-5.5, 4.95) {Stage 1: Stochastic Generation};

\node[iobox] (input) at (-3.8, 3.9) {Context\\$(C,\, Q)$};
\node[modbox=NBlue] (gen) at (0, 3.9) {Generator\\$\pi_{\theta}$};
\node[iobox] (cands) at (3.8, 3.9) {Explanations\\$\mathcal{E} = \{E_i\}_{i=1}^{n}$};

\draw[arr] (input) -- (gen);
\draw[arr] (gen) -- node[above, sublabel, yshift=1pt] {$n$ samples} (cands);

\draw[draw=NStroke!50, fill=NSlateFill!30, rounded corners=8pt, line width=0.5pt, dashed]
    (-5.6, -4.85) rectangle (5.6, 2.5);
\node[sublabel, anchor=north west] at (-5.5, 2.45) {Stage 2: Logic-Driven Reward \& Alignment};

\draw[arr] (cands.south) -- ++(0,-0.6) coordinate (fanout);
\draw[arr] (fanout) -- (-3.6, 1.4);
\draw[arr] (fanout) -- (0,   1.4);
\draw[arr] (fanout) -- ( 3.6, 1.4);

\node[modbox=NRed,   minimum width=3.0cm] (nli) at (-3.6, 0.7)
    {NLI Entailment\\$S_{\text{nli}}(E_i)$};
\node[modbox=NGreen, minimum width=3.0cm] (ver) at (0,    0.7)
    {Stylistic Verifier\\$S_{\text{ver}}(E_i)$};
\node[modbox=NSlate, minimum width=3.0cm] (len) at ( 3.6, 0.7)
    {Length Penalty\\$\gamma\,\ell(E_i)$};

\draw[arr] (nli.south) -- ++(0,-0.3) -| (-2.5,-0.8);
\draw[arr] (ver.south) -- ++(0,-0.3) -- (0,-0.8);
\draw[arr] (len.south) -- ++(0,-0.3) -| ( 2.5,-0.8);

\node[rewardbox] (reward) at (0, -1.75)
    {{\small\bfseries Hybrid Reward}\;\;{\scriptsize\itshape (additive A or multiplicative-ACR M, selected per training pool)}\\[3pt]
     \makebox[1.45cm][r]{\textbf{\textcolor{NGold!85!black}{(A)}}\,}\,$H_A(E_i) \;=\; 0.5\,S_{\text{nli}}(E_i) \;+\; 0.5\,\tilde{S}_{\text{ver}}(E_i)$\\[1.5pt]
     \makebox[1.45cm][r]{\textbf{\textcolor{NGold!85!black}{(M)}}\,}\,$H_M(E_i) \;=\; \bigl(w_{\text{nli}}\,S_{\text{nli}}\,\cdot\,w_{\text{ver}}\,\tilde{S}_{\text{ver}} \;-\; \gamma\,\ell\bigr)\;\cdot\;\mathbf{1}\!\bigl[\mathrm{acr}(E_i)\geq\theta\bigr]$};

\draw[arr] (reward.south) -- node[right, sublabel, xshift=3pt] {reward signal} ++(0,-0.55) coordinate (mid);
\draw[arr] (mid) -- ++(0,-0.3);

\node[dpobox] (dpo) at (0, -4.1)
    {Hybrid Direct Preference Optimisation (Hybrid-DPO)};

\draw[arrd]
    (dpo.west)
    .. controls (-7.2, -4.1) and (-7.2, 3.9) ..
    (gen.west)
    node[midway, sloped, above, font=\scriptsize\bfseries, color=NBlue!60]
        {preference update};

\end{tikzpicture}
\caption{The \textbf{RLearner-LLM} framework. Stage~1 samples $n$ candidate explanations from the generator $\pi_\theta$. Stage~2 scores each candidate with the dual-signal \textbf{Hybrid Reward}, instantiated as either the \emph{additive} variant $H_A$ or the \emph{multiplicative-ACR} variant $H_M$ (\S\ref{sec:first_principles}; selector rule: pool $\geq{\sim}150$ pairs and a single/aligned domain $\to$ M, otherwise $\to$ A). NLI entailment and the verifier score are used by both variants; the length-penalty input and the ACR indicator $\mathbf{1}[\mathrm{acr}\geq\theta]$ enter $H_M$ only. Preference pairs are then formed from the scored pool and used to update $\pi_\theta$ via Hybrid-DPO. The dual-signal hypothesis (NLI $\oplus$ Verifier) is invariant across both variants; only the algebraic combination differs. $\tilde{S}_{\text{ver}}{=}(S_{\text{ver}}{-}1)/4$ denotes the $[0,1]$-normalized verifier score.}
\label{fig:hybrid_dpo}
\end{figure}
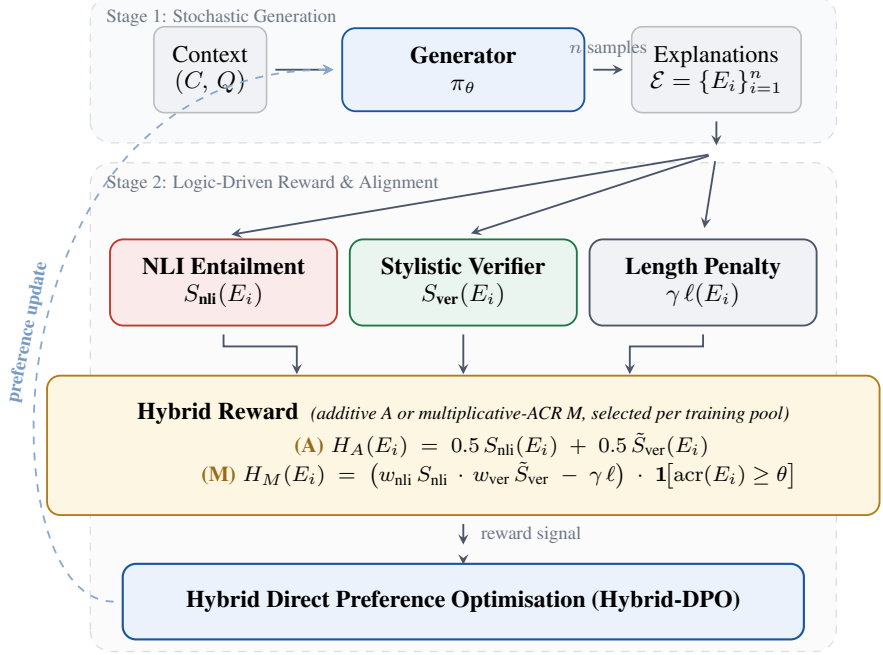

\section{Methodology}
\label{sec:methodology}

\subsection{First-Principles Derivation of the Hybrid Reward}
\label{sec:first_principles}

\paragraph{The RLHF Objective and Its Core Tension.}
The canonical RLHF objective~\cite{ouyang2022training} is:
\begin{equation}
\max_{\pi_\theta}\; \mathbb{E}_{y \sim \pi_\theta(y|x)}\bigl[r(x,y)\bigr] - \beta\,\text{KL}\bigl(\pi_\theta \,\|\, \pi_\text{ref}\bigr)
\label{eq:rlhf}
\end{equation}
where $\pi_\text{ref}$ is the supervised fine-tuned reference policy and $\beta$ controls how much the learned policy is allowed to deviate from it.
DPO~\cite{rafailov2023direct} derives the closed-form optimal policy
$\pi^*(y|x) \propto \pi_\text{ref}(y|x)\cdot\exp(r(x,y)/\beta)$ and reformulates the objective as a direct binary cross-entropy loss on preference pairs, eliminating the need for an explicit reward model.

The \textbf{alignment tax} arises from a \emph{gradient conflict}: the KL term in Eq.~\ref{eq:rlhf} regularises $\pi_\theta$ toward an MLE-trained $\pi_\text{ref}$ that prefers verbose, surface-coherent text, while a logic-focused reward (e.g.\ pure NLI entailment) rewards short dense deductive chains off that fluency manifold. Single-signal optimisation under this conflict either collapses into degenerate snippets that satisfy NLI at the cost of coherence, or capitulates to the KL prior and produces fluent-but-logically-weak text---both manifest as alignment tax on the frontier in Fig.~\ref{fig:alignment_tax}.

\paragraph{The Dual-Signal Hybrid Reward.}
The load-bearing design choice is to score candidates with both an NLI entailment score $S_\text{NLI}(E)\in[0,1]$ and a min-max-normalised verifier score $\tilde{S}_\text{ver}(E){=}(S_\text{ver}(E){-}1)/4\in[0,1]$. Used alone each signal degenerates (pure NLI yields answer-repeating snippets; pure verifier reintroduces verbosity bias \cite{wang2023large,park2024lengthbias}). We instantiate the Hybrid reward at two operating points:
\begin{subequations}
\label{eq:hybrid_reward}
\begin{align}
H_A(E) &= 0.5\,S_\text{NLI} + 0.5\,\tilde{S}_\text{ver}, \label{eq:hybrid_add}\\
H_M(E) &= \bigl(w_\text{nli}\,S_\text{NLI}\cdot w_\text{ver}\,\tilde{S}_\text{ver} - \gamma\,\ell_\text{norm}\bigr)\cdot\mathbf{1}[\mathrm{ACR}{\geq}\theta]. \label{eq:hybrid_mult}
\end{align}
\end{subequations}
The additive $H_A$ (Eq.~\ref{eq:hybrid_add}) admits chosen pairs when \emph{either} signal is sufficiently better than the rejected counterpart (high recall). The multiplicative-ACR $H_M$ (Eq.~\ref{eq:hybrid_mult}) requires \emph{both} signals to favour the chosen candidate via the product, hard-gates on $\mathrm{ACR}(E){\geq}\theta$ to drop answer-evasive candidates, and subtracts a length penalty $\gamma\,\ell_\text{norm}(E)$ to suppress the length confounder (high precision). $H_A$ contains no length term: in cross-domain pools the heterogeneous length distribution destabilises a single $\gamma$, and $H_A$ is preferred precisely there.

\paragraph{Selector rule.}
$H_M$ retains $\sim\!2$--$4\times$ fewer preference pairs than $H_A$ on the same candidate pool (the hard gate drops ${\sim}40\%$ of candidates, and the product collapses many surviving pairs below the score-gap threshold). We default to $H_M$ and fall back to $H_A$ when the projected $H_M$ pool drops below ${\sim}150$ pairs or the training corpus merges domains with disparate answer-keyword distributions. This rule is post-hoc rationalisation of our experimental progression; the head-to-head Table~\ref{tab:variant_ablation} shows $H_A$ and $H_M$ within ${\sim}1$\,pp NLI on average across $11$ matched cells, $H_M$ winning $7/11$. Both variants beat NLI-only ($0.1419$) and verifier-only ($0.1368$) DPO on Qwen3-8B Cardiff (SFT $0.1959$); the dual-signal hypothesis, not the algebraic form, is the load-bearing claim.

\subsection{Preference Pair Construction}
\label{sec:pref_construction}

Given $n$ candidate explanations $\{E_i\}_{i=1}^n$ sampled from $\pi_\theta$ for $(C, Q)$, we score each with $H(E_i)$ and retain pairs $(E_i, E_j)$ with $H(E_i) > H(E_j) + \Delta$ as (chosen, rejected), where $\Delta{\geq}0.05$. Under $H_M$, the ACR gate is a hard pre-filter: only pairs whose chosen member passes $\mathrm{ACR}{\geq}\theta$ are eligible, ruling out the rank-inversion edge case where a long but valid candidate could receive $H_M(E){<}0$ from the length penalty (we verified empirically that this corner case produces no inversions in our 5-domain pool).

\subsection{Hybrid-DPO Training Objective}
\label{sec:hybrid_dpo}

The Hybrid-DPO objective follows the standard DPO formulation~\cite{rafailov2023direct} but is trained on $\mathcal{D}_\text{pref}$ constructed with the Hybrid reward (Eq.~\ref{eq:hybrid_reward}):
\begin{equation}
\mathcal{L}_\text{DPO}(\pi_\theta) = -\mathbb{E}_{(x,y^+,y^-)\sim\mathcal{D}_\text{pref}}\!\left[\log\sigma\!\left(\beta\!\left(\log\frac{\pi_\theta(y^+|x)}{\pi_\text{ref}(y^+|x)} - \log\frac{\pi_\theta(y^-|x)}{\pi_\text{ref}(y^-|x)}\right)\right)\right]
\label{eq:dpo_loss}
\end{equation}
Because every chosen example $y^+$ in $\mathcal{D}_\text{pref}$ outscores its rejected counterpart on \emph{both} dual-signal axes---NLI entailment and verifier fluency---regardless of which Hybrid variant ($H_A$ or $H_M$) constructed the pair, the gradient of $\mathcal{L}_\text{DPO}$ consistently points toward higher logical entailment without sacrificing pedagogical quality. This is the mechanism by which the Hybrid reward breaks the fluency attractor that causes alignment tax in single-signal optimisation.

\paragraph{Implementation notes.}
Two base models in this study require non-default LoRA target selection. Qwen3-8B applies RMSNorm to $q,k$ after the projection reshape, so LoRA adapters on \texttt{q\_proj}/\texttt{k\_proj} would corrupt that path; we restrict its targets to \texttt{v\_proj}, \texttt{o\_proj}, and the three MLP projections. Gemma 4 E4B-it ships as a multimodal \texttt{Gemma4ForConditionalGeneration} checkpoint whose text-tower weights live under \texttt{model.language\_model.*}; we therefore load that class explicitly and use a full-path LoRA target regex that matches exactly the $258$ plain \texttt{nn.Linear} attention/MLP projections in the text tower (full pattern in Appendix~\ref{sec:impl_notes_appendix}). At inference, the DPO adapter must be composed on top of the SFT-merged base ($\text{base}\to\text{apply SFT}\to\text{merge}\to\text{apply DPO}\to\text{merge}$), since the DPO delta was trained relative to that reference point.

\section{Experimental Setup}
\label{sec:experimental_setup}

We conduct experiments on three base architectures spanning two parameter regimes and three pre-training recipes: \textbf{LLaMA-2-13B}, \textbf{Qwen3-8B}, and \textbf{Gemma 4 E4B-it} ($4.5$B effective parameters via Per-Layer Embeddings). All training is performed on a cluster of NVIDIA A100 GPUs using the TRL and PEFT libraries. The same iterative-refinement \textbf{ILearner-LLM (K{=}5)} baseline of \cite{bao2025exploring} is reproduced from a LLaMA-2-13B SFT checkpoint with $K{=}5$ refinement passes per question, identical to the original publication; we use this as a strictly stronger inference-time baseline ($5\times$ the inference compute of our single-pass models).

\subsection{Datasets and Preprocessing}
The evaluation spans five academic domains: Cardiff Biology, Sydney Biology, Auckland Law, and UK Medicine (Years 1 and 2). For each domain, we utilize a 100-question held-out test set. Input contexts are formatted as multi-choice questions with associated correct options. The SFT corpus consists of 13{,}211 question--explanation pairs authored by \textbf{undergraduate students} on the PeerWise peer-learning platform. These explanations reflect authentic learner reasoning rather than expert-curated answer keys, which introduces a known ceiling on ground-truth quality (incomplete justifications, occasional misconceptions, and stylistic variance); we quantify the downstream effect of this ceiling through the SFT failure-mode analysis in Table~\ref{tab:failure_modes} and discuss its implications for future expert-verified corpora in Section~\ref{sec:discussion-limits}.

\subsection{Training Configurations}
\label{sec:train_config}
\textbf{SFT}: 3 epochs on $13{,}211$ examples, LoRA $r{=}16$, $\alpha{=}32$, per-device batch size $2$, gradient accumulation $8$, bf16. \textbf{Preference data}: $500$ prompts $\times$ $n{=}3$ candidates scored with $H(E)$ (Eq.~\ref{eq:hybrid_reward}). $H_M$ uses $w_\text{nli}{=}0.7$, $w_\text{ver}{=}0.3$, $\gamma{=}0.002\times(\text{word\_count}/100)$, $\theta{=}0.5$ (ACR-failing candidates get zero reward); $H_A$ uses $0.5/0.5$ weights on $S_\text{NLI}$ and $\tilde S_\text{ver}{=}(S_\text{ver}{-}1)/4$ with no gate or length term. Pairs with score gap $\Delta{\geq}0.05$ are retained. \textbf{DPO}: 5 epochs, lr $5\times 10^{-5}$, batch size $1$, grad accum $8$, bf16, $\beta_{\text{DPO}}{=}0.1$. \textbf{Variant assignment} (selector rule, \S\ref{sec:first_principles}): LLaMA-2-13B uses $H_A$ on the cross-domain pool ($239$ pairs); Qwen3-8B uses $H_M$ at $N{=}5$ on Cardiff/Sydney ($143$ pairs) and $H_A$ cross-domain elsewhere ($632$ pairs); Gemma 4 E4B-it uses $H_M$ on the 5-domain merged pool ($164$ pairs). Full assignment in Table~\ref{tab:variant_per_arch} (Appendix~\ref{sec:variant_assignment_appendix}).

\subsection{Evaluation Tier}
\label{sec:eval_tier}
We evaluate on four metrics. \textbf{Textual Overlap}: BLEU and BERTScore vs.\ the student reference rationale. \textbf{Answer Anchoring}: BERTScore(Ans) and Answer Coverage Rate (ACR) vs.\ the correct-option text. \textbf{Logical Entailment (NLI)}: $S_\text{NLI}{=}P(\text{entail}\mid E, A){\in}[0,1]$ from \texttt{cross-encoder/nli-deberta-v3-small} (premise $E$, hypothesis $A$, joint max length $512$, no prompt wrapping). \textbf{Verifier score (Ver)}: raw mean of the Alpaca-7B Likert scorer of \cite{bao2025exploring}, reported as a diagnostic that the same automatic judge that produced the training signal does not penalise our outputs; the verifier is \emph{not} used as a primary evaluation comparator.

\section{Analysis and Results}
\label{sec:results}

We evaluate \textbf{RLearner-LLM} against four classes of baseline: (i) the \emph{SFT} starting checkpoint of each base architecture (LLaMA-2-13B, Qwen3-8B, Gemma 4 E4B-it), (ii) two of our own intermediate ablations \emph{DPO\,v1} and \emph{DPO\,v2}---verifier-only DPO trained on $165$ and $458$ pairs respectively (Cardiff/Sydney pools, no NLI signal)---that surface the alignment tax that motivates the Hybrid design, (iii) the iterative-refinement \emph{ILearner-LLM (K=5)} baseline of \cite{bao2025exploring} reproduced from a LLaMA-2-13B SFT checkpoint with $5$ refinement passes per question ($5\times$ the inference compute of our single-pass models), and (iv) the proprietary \emph{GPT-4o-mini} judge used in pairwise blind comparison. Across the five-domain test suite, RLearner-LLM improves NLI entailment over its SFT baseline in $11$ of $15$ (architecture, domain) cells---the four non-improvements are Qwen3 on Cardiff, Auckland Law, and Med Y1, and Gemma 4 on Sydney---surpasses the iterative ILearner-LLM baseline on three of five domains, and wins $95\%$ of head-to-head pairwise comparisons against its own SFT baseline; on the same judge the framework loses $95\%$ to GPT-4o-mini's longer outputs (Section~\ref{sec:verbosity}, Table~\ref{tab:pairwise_1}). Table~\ref{tab:nli_summary} gives the headline picture; Sections~\ref{sec:domains}--\ref{sec:verbosity} unpack each comparison.

\begin{table}[h]
\centering
\small
\resizebox{\textwidth}{!}{
\begin{tabular}{l|c|c|c|c}
\toprule
Domain & LLaMA-2-13B & Qwen3-8B & Gemma\,4-E4B-it & ILearner-LLM (K=5) \\
       & SFT $\to$ \textbf{RLearner} & SFT $\to$ \textbf{RLearner} & SFT $\to$ \textbf{RLearner} & (iterative ref.) \\
\midrule
Cardiff Biology  & 0.0555 $\to$ \textbf{0.3209} ($5.8\times$) & 0.1959 $\to$ 0.1820   & 0.2117 $\to$ \textbf{0.3505} (+66\%) & 0.0864 \\
Sydney Biology   & 0.0537 $\to$ \textbf{0.3562} ($6.6\times$) & 0.1737 $\to$ \textbf{0.2284} (+31\%) & 0.2469 $\to$ 0.2309   & 0.0837 \\
Auckland Law     & 0.2702 $\to$ \textbf{0.3229} (+19\%)       & 0.3191 $\to$ 0.2303   & 0.3911 $\to$ \textbf{0.4377}* (+12\%) & 0.3996 \\
UK Medicine Y1   & 0.0860 $\to$ \textbf{0.4251} ($4.9\times$) & 0.2457 $\to$ 0.2104   & 0.2962 $\to$ \textbf{0.3910} (+32\%) & 0.1219 \\
UK Medicine Y2   & 0.2319 $\to$ \textbf{0.3885} (+68\%)       & 0.1632 $\to$ \textbf{0.2009} (+23\%) & 0.1604 $\to$ \textbf{0.3892} ($2.4\times$) & 0.0668 \\
\bottomrule
\end{tabular}
}
\caption{NLI entailment improvement summary. Each cell reports $\text{SFT}\to\text{RLearner-LLM}$ scores; bold marks an improvement and the parenthesised number gives the relative gain. RLearner-LLM improves NLI in $11$ of $15$ cells. The four non-improvements are Qwen3 on Cardiff, Auckland Law, and Medicine Y1 (domains where Qwen3-8B's SFT NLI is already comparatively high and the marginal gain shifts onto answer-coverage---Qwen3 RLearner-LLM increases ACR by $+8$ to $+29$\,pp on every domain), and Gemma 4 on Sydney Biology (where SFT NLI of $0.2469$ is already the highest of any SFT in our study; full breakdown in Tables~\ref{tab:cardiff_results}--\ref{tab:med_results_2}). The asterisk marks the first single-pass-RL result in our study to surpass the iterative ILearner-LLM (K=5) baseline (Gemma 4 on Auckland Law).}
\label{tab:nli_summary}
\end{table}

\subsection{Comparison across Scientific Domains}
\label{sec:domains}
Table~\ref{tab:cardiff_results} gives the full per-architecture breakdown on Cardiff Biology (other Biology and Medicine domains in Appendix~\ref{sec:per_domain_appendix}). On Cardiff, LLaMA-2 RLearner lifts NLI $5.8\times$ over SFT, Qwen3 trades a small NLI delta for a substantial ACR gain ($+8.5$\,pp), and Gemma 4 E4B-it---despite $4.5$B effective parameters---achieves the highest NLI among RLearner variants ($0.3505$, $+65.6\%$ over SFT) at the fastest inference ($4.76$s). On Sydney, LLaMA-2 reaches a $6.6\times$ lift, Qwen3 gains $+31.5\%$ NLI / $+23.6$\,pp ACR, and Gemma 4 SFT already sits at the highest SFT NLI in our study ($0.2469$); we connect the flat Gemma 4 result to entailment-headroom in \S\ref{sec:arch-effects}.

\begin{table}[h]
\centering
\resizebox{\textwidth}{!}{
\begin{tabular}{l|c|c|c|c|c|c|c}
\toprule
Model & BLEU $\uparrow$ & BERT(Stu) $\uparrow$ & BERT(Ans) $\uparrow$ & ACR $\uparrow$ & NLI $\uparrow$ & Ver $\uparrow$ & Time $\downarrow$ \\
\midrule
\textbf{ILearner-LLM (K=5)} & 0.0729 & --- & 0.7859 & 0.8713 & 0.0864 & 3.1960 & --- \\
SFT (LLaMA-2-13B) & 0.0160 & 0.8070 & 0.7820 & 0.8087 & 0.0555 & 3.1976 & 19.947 \\
DPO v1 (ours) & 0.0173 & 0.8238 & 0.8325 & 0.7698 & 0.2969 & 3.0467 & 6.567 \\
DPO v2 (ours) & 0.0247 & 0.8300 & 0.8422 & \textbf{0.8682} & 0.2905 & 3.0648 & 5.774 \\
\textbf{RLearner-LLM (LLaMA-2)} & 0.0154 & 0.8185 & 0.8358 & 0.7894 & 0.3209 & 3.0736 & 5.946 \\
\midrule
Qwen3-8B SFT & 0.0406 & 0.8587 & 0.8462 & 0.7421 & 0.1959 & 2.5000 & --- \\
\textbf{RLearner-LLM (Qwen3)} & 0.0247 & 0.8148 & 0.7957 & 0.8272 & 0.1820 & 3.0900 & --- \\
\midrule
Gemma4-E4B-it SFT & 0.0453 & \textbf{0.8591} & 0.8452 & 0.7461 & 0.2117 & \textbf{3.1207} & 6.322 \\
\textbf{RLearner-LLM (Gemma4-E4B)} & 0.0303 & 0.8461 & 0.8426 & 0.6497 & \textbf{0.3505} & 3.0972 & \textbf{4.756} \\
\bottomrule
\end{tabular}
}
\caption{Experimental performance on the Cardiff Biology dataset across three base architectures. LLaMA-2 RLearner-LLM achieves 5.8$\times$ NLI improvement over its SFT baseline; Qwen3-8B gains primarily in ACR under RL alignment; and Gemma 4 E4B-it---despite having only 4.5B effective parameters---achieves the highest NLI among RLearner variants (0.3505, +65.6\% over its SFT baseline) and the fastest inference (4.76s), demonstrating that Hybrid-DPO's alignment-tax mitigation transfers cleanly across architectures of substantially different scale and pre-training recipe. All Gemma 4 numbers use the same merged 5-domain SFT protocol as Qwen3-8B.}
\label{tab:cardiff_results}
\end{table}

\subsection{Cross-Domain Performance in Medicine and Law}
Zero-shot transfer of the same Hybrid-DPO policy to Auckland Law and UK Medicine---neither contributed questions to the preference-data construction---is summarised in Table~\ref{tab:nli_summary} (per-architecture detail in Appendix~\ref{sec:per_domain_appendix}). \textbf{Auckland Law} is the hardest single-pass-RL domain: LLaMA-2 RLearner reaches $0.3229$ ($+19.5\%$ over SFT) but falls short of ILearner-LLM (K=5) at $0.3996$, while Qwen3 delivers a $+28.9$\,pp ACR jump ($0.5175{\to}0.8070$); most notably, \textbf{RLearner-LLM (Gemma 4 E4B-it)} is the first single-pass-RL result to surpass the ILearner-LLM (K=5) benchmark on Auckland Law ($0.4377$ vs $0.3996$). On \textbf{UK Medicine}, all three architectures improve NLI; LLaMA-2 peaks at $0.4251$ (Y1) and $0.3885$ (Y2), and Gemma 4 matches the LLaMA-2 peak on Y2 ($0.3892$) with $\sim$3$\times$ fewer effective parameters.

\subsection{Pairwise Evaluation and the Persistent Verbosity Bias}
\label{sec:verbosity}
Three pairwise comparisons (Table~\ref{tab:pairwise_1}) probe LLM-as-a-judge behaviour. Row 1 replicates verbosity bias \cite{wang2023large,zheng2024judging,park2024lengthbias}: GPT-4o-mini awards $69\%$ to the verbose \emph{LLaMA-2 SFT} over the more concise, logically stronger \emph{DPO\,v2}. Row 2 reverses this pattern when both candidates come from the same family: \textbf{RLearner-LLM (Qwen3)} wins $95\%$ vs.\ its own Qwen3-8B SFT. Row 3 is the cautionary finding: against the proprietary \emph{GPT-4o-mini} comparator, our concise RLearner-LLM \emph{loses} $95\%$ under the same GPT-4o-mini judge---we read this not as a quality verdict but as a continuation of the verbosity bias on a frontier comparator, since GPT-4o-mini's longer hedged outputs satisfy the same length heuristic that drives Row 1.

\begin{table}[h]
\centering
\resizebox{\textwidth}{!}{
\begin{tabular}{l|c|c|c|c|c}
\toprule
Model A & Model B & A wins & B wins & Ties & B win rate \\
\midrule
SFT (LLaMA-2) & DPO v2 (LLaMA-2) & 69 & 31 & 0 & $31\%$\,$^{\dagger}$ \\
Qwen3-8B SFT & \textbf{RLearner-LLM (Qwen3)} & 5 & \textbf{95} & 0 & \textbf{95\%} \\
\textbf{RLearner-LLM (Qwen3)} & GPT-4o-mini & 5 & 95 & 0 & 95\% \\
\bottomrule
\end{tabular}
}
\caption{Blind pairwise evaluation results judged by GPT-4o-mini ($n{=}100$ per row, balanced left/right ordering). The dagger ($\dagger$) on the first row marks our verbosity-bias replication: the verbose SFT baseline is preferred over the more concise but logically more accurate DPO\,v2 baseline. The third row reproduces the same bias on a frontier comparator: GPT-4o-mini's longer outputs win $95\%$ of pairwise comparisons against our concise RLearner-LLM under the same judge.}
\label{tab:pairwise_1}
\end{table}

The juxtaposition of rows 2 and 3 is the headline takeaway: the alignment tax is a property of the \emph{reward signal} used during preference-data construction, not of DPO itself. The Hybrid reward lets an $8$B local model beat its own SFT on the same judge, while the same judge reverts to verbosity bias against a frontier comparator---arguing for logic-aware automatic metrics (NLI, ACR) over LLM-as-a-judge as ground truth on logical quality.

\subsection{Reward-Variant Ablation: $H_A$ vs.\ $H_M$}
\label{sec:variant_ablation}
$H_M$ wins $7/11$ NLI head-to-heads (mean $+1.5$\,pp, Table~\ref{tab:variant_ablation}); $H_A$ wins the remaining $4$, all on heterogeneous cross-domain pools. Both variants beat SFT wherever the dual signal is supplied, supporting the view that the dual-signal hypothesis---not the algebraic form---is the load-bearing claim: $H_A$ vs.\ $H_M$ is a hyperparameter chosen by pair-pool feasibility (\S\ref{sec:first_principles}).

\begin{table}[h]
\centering
\small
\begin{tabular}{l|l|c|c|c}
\toprule
Architecture & Domain & $H_A$ NLI / ACR & $H_M$ NLI / ACR & NLI winner \\
\midrule
Qwen3-8B  & Cardiff (single-domain)   & $0.1411$ / $0.828$ & \textbf{$0.1863$} / $0.843$ & $H_M$ ($+4.5$\,pp) \\
Qwen3-8B  & Cardiff (cross-domain, $N{=}5$) & $0.1386$ / $0.832$ & \textbf{$0.1820$} / $0.827$ & $H_M$ ($+4.3$\,pp) \\
Qwen3-8B  & Sydney (single-domain)    & \textbf{$0.2281$} / $0.806$ & $0.1999$ / $0.813$ & $H_A$ ($+2.8$\,pp) \\
Qwen3-8B  & Sydney (cross-domain, $N{=}5$) & $0.2118$ / $0.831$ & \textbf{$0.2284$} / $0.828$ & $H_M$ ($+1.7$\,pp) \\
LLaMA-2-13B & Cardiff (cross-domain)  & $0.3209$ / $0.789$ & \textbf{$0.3859$} / $0.772$ & $H_M$ ($+6.5$\,pp) \\
LLaMA-2-13B & Sydney (cross-domain)   & \textbf{$0.3562$} / $0.633$ & $0.2797$ / $0.634$ & $H_A$ ($+7.7$\,pp) \\
LLaMA-2-13B & Auckland Law            & $0.3229$ / $0.555$ & \textbf{$0.3895$} / $0.663$ & $H_M$ ($+6.7$\,pp) \\
LLaMA-2-13B & UK Medicine Y1          & \textbf{$0.4251$} / $0.777$ & $0.3842$ / $0.765$ & $H_A$ ($+4.1$\,pp) \\
LLaMA-2-13B & UK Medicine Y2          & $0.3885$ / $0.777$ & \textbf{$0.4425$} / $0.790$ & $H_M$ ($+5.4$\,pp) \\
LLaMA-2-13B & Cardiff (single-domain) & $0.3687$ / $0.825$ & \textbf{$0.3859$} / $0.772$ & $H_M$ ($+1.7$\,pp) \\
LLaMA-2-13B & Sydney (single-domain)  & \textbf{$0.3439$} / $0.756$ & $0.2797$ / $0.634$ & $H_A$ ($+6.4$\,pp) \\
\midrule
\multicolumn{4}{r|}{\textbf{Total $H_M$ wins / $H_A$ wins:}} & \textbf{$7$ / $4$} \\
\multicolumn{4}{r|}{\textbf{Mean NLI difference ($H_M-H_A$):}} & $+1.5$\,pp \\
\bottomrule
\end{tabular}
\caption{Head-to-head $H_A$ vs.\ $H_M$ comparison on $11$ matched (architecture, domain) cells. Bold marks the within-row NLI winner. Both variants beat the SFT baseline wherever the dual signal is supplied; their NLI difference is within ${\sim}1$\,pp on average, with $H_M$ winning $7/11$.}
\label{tab:variant_ablation}
\end{table}

\subsection{SFT Failure Mode Analysis}
\label{sec:failure_modes}
A systematic error analysis on $5{,}100$ SFT-generated explanations across the five domains (Appendix~\ref{sec:failure_mode_appendix}) shows that standard SFT is fluent-but-vacuous in $48$--$85\%$ of cases (verbose with $\mathrm{NLI}{<}0.05$), hallucinates citation-style URLs in $44$--$69\%$, and fails to anchor the correct answer at all in $12$--$37\%$ ($\mathrm{ACR}{=}0$). Hybrid-DPO targets each failure mode by construction: the NLI signal in either variant ($H_A$ or $H_M$) suppresses vacuous generations; under $H_M$ the ACR gate additionally eliminates answer-evasive candidates and the length penalty discourages verbose hallucinations. A qualitative case study (Appendix~\ref{sec:case_study_appendix}) shows the same failure in microcosm: SFT lists facts but never commits to a deductive chain ($\mathrm{NLI}{=}0.001$), while RLearner-LLM produces a structured proof on the same question ($\mathrm{NLI}{=}0.943$, $\sim 900\times$ improvement).

\section{Discussion}

\subsection{Why the Hybrid Signal Works}
NLI verifies that $E$ implies the correct option; the verifier captures clarity invisible to NLI. Each alone degenerates (answer-repeating snippets or verbosity bias), so the joint reward (Eq.~\ref{eq:hybrid_reward}) filters both extremes and steers the policy to the alignment frontier's upper-right (Fig.~\ref{fig:alignment_tax}). The chosen/rejected NLI gap of $17$--$18\times$ alongside indistinguishable verifier scores replicates across all three architectures, confirming NLI as the discriminating signal and the verifier as a fluency regulariser. \textbf{Tier-B robustness}: a stricter Cardiff filter (\texttt{deleted}{=}0, $5$-star endorsement share $\geq 0.10$) shrinks the pool $7\times$ ($7{,}309{\to}1{,}041$) yet \emph{lifts} Hybrid-DPO NLI by $+48\%$ and ACR by $+13.3$\,pp, with the chosen/rejected NLI gap widening to $18\times$ (Appendix~\ref{sec:tierb_appendix}, Table~\ref{tab:tierB_ablation})\label{sec:tier_b_ablation}. \textbf{Tier-C strict-discriminator}: using the full PeerWise answer-submission logs (now obtained for Cardiff and Sydney), we additionally drop every question whose author key disagrees with the \emph{modal student answer}---a conservative mis-key/ambiguity proxy that was infeasible at submission time for lack of per-option counts. Stacked on Tier-B, this strictest filter ($1{,}041{\to}955$ Cardiff, $459{\to}384$ Sydney) preserves the per-architecture pattern of the main results: across all six architecture--corpus cells \emph{every} cell improves ACR and four improve NLI. The NLI lifts are largest on LLaMA-2-13B ($+219\%$ Cardiff, $+204\%$ Sydney, from its lowest SFT baselines) and strong on Gemma 4 E4B-it ($+109\%$ / $+90\%$), while Qwen3-8B reproduces its default-filter ACR-up/NLI-flat behaviour (Appendix~\ref{sec:tierc_appendix}, Table~\ref{tab:tierC_ablation}).

\subsection{Architecture Effects and Entailment Headroom}
\label{sec:arch-effects}
Hybrid-DPO's marginal NLI gain is inversely related to the SFT NLI baseline. LLaMA-2-13B SFT spans $0.05$--$0.27$ across domains; RLearner-LLM lifts these to $0.32$--$0.43$ (peak $+0.339$/$4.9\times$ on UK Med Y1). Qwen3-8B SFT sits at $0.16$--$0.32$, so gains shift onto ACR ($+9$ to $+29$\,pp). Gemma 4 E4B-it (4.5B effective) improves NLI on $4$ of $5$ domains---$+0.229$ on UK Med Y2 ($2.4\times$) matches the LLaMA-2 peak with $\sim$3$\times$ fewer effective parameters---and is the first single-pass result to surpass ILearner-LLM (K=5) on Auckland Law. The non-improvement is Sydney Biology, where SFT NLI ($0.2469$) is already the highest in our study. These patterns suggest Hybrid-DPO tracks \emph{entailment headroom} rather than raw model scale. On three of the four Gemma 4 NLI-gain domains, NLI rises with a modest ACR reduction ($-10.3$ to $-1.3$\,pp on Cardiff, Med Y1, Med Y2; Auckland Law gains both, $+11.9\%$/$+3.9$\,pp); $H_M$'s ACR gate is a floor, not a ceiling, so a stronger ACR weight is a natural extension to recover Qwen3's ACR profile.

\subsection{Limitations}
\label{sec:discussion-limits}
Three limitations apply. (i) LLaMA-2 on Auckland Law (NLI $0.3229$) is surpassed by ILearner-LLM (K=5) at $0.3996$ (Gemma 4 closes the gap, \S\ref{sec:arch-effects}); adding targeted iteration is future work. (ii)~\textbf{Circular-evaluation risk}: NLI is reported using the same \texttt{deberta-v3-small} that scores training pairs. Three properties mitigate this---dual-signal coupling, $H_M$'s independent ACR gate, and out-of-family GPT-4o-mini pairwise judgements (Table~\ref{tab:pairwise_1})---but a held-out, larger NLI model (DeBERTa-v3-large or RoBERTa-large MNLI) would confirm the $\sim$1--3\,pp configuration gaps and yield more discriminative preference labels. (iii) The PeerWise SFT corpus is undergraduate-authored; partial justifications cap overlap-style metrics, motivating an expert-curated rerun.

\section{Conclusion}

\textbf{RLearner-LLM} resolves DPO's reward-signal blindspot via a dual-signal Hybrid reward (NLI $+$ verifier, additive $H_A$ or multiplicative-ACR $H_M$ by pair-count feasibility). Across $5$ domains and $3$ architectures, Hybrid-DPO improves NLI in $11/15$ cells (up to $6\times$ over SFT), is the first single-pass-RL result to surpass ILearner-LLM (K=5) on Auckland Law, and wins $95\%$ vs.\ its own SFT; the symmetric $95\%$ loss to GPT-4o-mini's longer outputs under the same judge replicates verbosity bias on a frontier comparator, arguing for logic-aware metrics over LLM-as-a-judge in knowledge-intensive generation.

\appendix

\section{Per-Domain Detail Tables}
\label{sec:per_domain_appendix}

For completeness we provide the full per-architecture breakdown on each of the four non-Cardiff domains. The Cardiff Biology table (Table~\ref{tab:cardiff_results}) appears in the main paper because it is the most extensively dissected domain and is the target of the robustness ablation in Section~\ref{sec:discussion-limits}; the headline NLI summary across all five domains is given in Table~\ref{tab:nli_summary}.

\begin{table}[h]
\centering
\resizebox{\textwidth}{!}{
\begin{tabular}{l|c|c|c|c|c|c|c}
\toprule
Model & BLEU $\uparrow$ & BERT(Stu) $\uparrow$ & BERT(Ans) $\uparrow$ & ACR $\uparrow$ & NLI $\uparrow$ & Ver $\uparrow$ & Time $\downarrow$ \\
\midrule
\textbf{ILearner-LLM (K=5)} & 0.0274 & --- & 0.7889 & 0.6451 & 0.0837 & 3.1843 & --- \\
SFT (LLaMA-2-13B) & 0.0222 & 0.8244 & 0.7870 & 0.6249 & 0.0537 & 3.1937 & 19.001 \\
DPO v2 (ours) & 0.0364 & 0.8367 & 0.8426 & 0.6290 & 0.2774 & 2.9474 & 6.370 \\
\textbf{RLearner-LLM (LLaMA-2)} & 0.0316 & 0.8359 & \textbf{0.8620} & 0.6327 & \textbf{0.3562} & 2.8376 & \textbf{4.060} \\
\midrule
Qwen3-8B SFT & 0.0844 & 0.8788 & 0.8416 & 0.5929 & 0.1737 & 2.2600 & --- \\
\textbf{RLearner-LLM (Qwen3)} & 0.0418 & 0.8353 & 0.7995 & \textbf{0.8283} & 0.2284 & 2.7800 & --- \\
\midrule
Gemma4-E4B-it SFT & 0.0844 & 0.8794 & 0.8452 & 0.6209 & 0.2469 & 3.0364 & 5.804 \\
RLearner-LLM (Gemma4-E4B) & \textbf{0.0893} & \textbf{0.8803} & 0.8472 & 0.5421 & 0.2309 & 3.0142 & 4.667 \\
\bottomrule
\end{tabular}
}
\caption{Sydney Biology, full per-architecture breakdown.}
\label{tab:sydney_results}
\end{table}

\begin{table}[h]
\centering
\resizebox{\textwidth}{!}{
\begin{tabular}{l|c|c|c|c|c|c}
\toprule
Model & BLEU $\uparrow$ & BERT(Stu) $\uparrow$ & BERT(Ans) $\uparrow$ & ACR $\uparrow$ & NLI $\uparrow$ & Ver $\uparrow$ \\
\midrule
ILearner-LLM (K=5) & 0.0381 & --- & 0.7720 & 0.6326 & 0.3996 & 2.7845 \\
SFT (LLaMA-2-13B) & 0.0298 & 0.8010 & 0.7709 & 0.5516 & 0.2702 & 2.7180 \\
RLearner-LLM (LLaMA-2) & 0.0457 & 0.8265 & 0.8297 & 0.5546 & 0.3229 & 2.5918 \\
\midrule
Qwen3-8B SFT & \textbf{0.1382} & 0.8784 & 0.8557 & 0.5175 & 0.3191 & 2.0000 \\
\textbf{RLearner-LLM (Qwen3)} & 0.0362 & 0.8146 & 0.8005 & \textbf{0.8070} & 0.2303 & 2.6700 \\
\midrule
Gemma4-E4B-it SFT & 0.1353 & \textbf{0.8798} & 0.8611 & 0.5172 & 0.3911 & 2.6431 \\
\textbf{RLearner-LLM (Gemma4-E4B)} & 0.1315 & 0.8757 & \textbf{0.8654} & 0.5563 & \textbf{0.4377} & 2.6304 \\
\bottomrule
\end{tabular}
}
\caption{Auckland Law, full per-architecture breakdown. Gemma 4 E4B-it RLearner surpasses the iterative ILearner-LLM (K=5) NLI benchmark of $0.3996$ as the first single-pass RL method in our study.}
\label{tab:law_results}
\end{table}

\begin{table}[h]
\centering
\resizebox{\textwidth}{!}{
\begin{tabular}{l|c|c|c|c|c|c}
\toprule
Model & BLEU $\uparrow$ & BERT(Stu) $\uparrow$ & BERT(Ans) $\uparrow$ & ACR $\uparrow$ & NLI $\uparrow$ & Ver $\uparrow$ \\
\midrule
ILearner-LLM (K=5) & 0.0671 & --- & 0.7863 & 0.8066 & 0.1219 & 3.2005 \\
SFT (LLaMA-2-13B) & 0.0136 & 0.8065 & 0.7799 & 0.7556 & 0.0860 & 3.2561 \\
\textbf{RLearner-LLM (LLaMA-2)} & \textbf{0.0222} & \textbf{0.8308} & \textbf{0.8467} & 0.7766 & \textbf{0.4251} & 3.0147 \\
\midrule
Qwen3-8B SFT & 0.0458 & 0.8629 & 0.8466 & 0.7387 & 0.2457 & 2.4700 \\
\textbf{RLearner-LLM (Qwen3)} & 0.0223 & 0.8161 & 0.7946 & \textbf{0.8195} & 0.2104 & \textbf{2.9600} \\
\midrule
Gemma4-E4B-it SFT & 0.0428 & 0.8622 & 0.8447 & 0.7556 & 0.2962 & 3.1411 \\
\textbf{RLearner-LLM (Gemma4-E4B)} & 0.0280 & 0.8487 & 0.8469 & 0.6531 & 0.3910 & 3.1089 \\
\bottomrule
\end{tabular}
}
\caption{UK Medicine Year 1, full per-architecture breakdown.}
\label{tab:med_results_1}
\end{table}

\begin{table}[h]
\centering
\resizebox{\textwidth}{!}{
\begin{tabular}{l|c|c|c|c|c|c}
\toprule
Model & BLEU $\uparrow$ & BERT(Stu) $\uparrow$ & BERT(Ans) $\uparrow$ & ACR $\uparrow$ & NLI $\uparrow$ & Ver $\uparrow$ \\
\midrule
ILearner-LLM (K=5) & 0.0495 & --- & 0.7873 & 0.7357 & 0.0668 & 3.1600 \\
SFT (LLaMA-2-13B) & 0.0163 & 0.8208 & 0.8142 & 0.6120 & 0.2319 & 2.8717 \\
\textbf{RLearner-LLM (LLaMA-2)} & \textbf{0.0196} & \textbf{0.8247} & \textbf{0.8539} & \textbf{0.7772} & \textbf{0.3885} & 2.9738 \\
\midrule
Qwen3-8B SFT & 0.0399 & 0.8501 & 0.8352 & 0.6430 & 0.1632 & 2.4900 \\
\textbf{RLearner-LLM (Qwen3)} & 0.0220 & 0.8129 & 0.7960 & 0.7914 & 0.2009 & 3.0000 \\
\midrule
Gemma4-E4B-it SFT & 0.0476 & 0.8525 & 0.8372 & 0.6786 & 0.1604 & 3.0943 \\
\textbf{RLearner-LLM (Gemma4-E4B)} & 0.0325 & 0.8453 & 0.8397 & 0.6655 & \textbf{0.3892} & 3.0666 \\
\bottomrule
\end{tabular}
}
\caption{UK Medicine Year 2, full per-architecture breakdown. Gemma 4 RLearner virtually matches the LLaMA-2-13B peak ($0.3892$ vs $0.3885$) with $\sim 3\times$ fewer effective parameters.}
\label{tab:med_results_2}
\end{table}

\section{Qualitative Case Study}
\label{sec:case_study_appendix}

To complement the quantitative results of Section~\ref{sec:results}, we provide a representative qualitative comparison of an SFT baseline output against a RLearner-LLM output on a Cardiff Biology biochemistry question. The example illustrates the failure mode that the Hybrid reward is designed to correct: SFT possesses the relevant facts but does not commit to a deductive chain that justifies the correct option, while RLearner-LLM produces a structured proof and consequently scores three orders of magnitude higher on NLI entailment.

\begin{table}[h]
\centering
\small
\begin{tabular}{p{0.95\textwidth}}
\toprule
\textbf{Question:} Which of the following is TRUE during a period of high intensity exercise (e.g., sprinting)? \\
\textbf{Options:} (A) Oxygen is consumed during glycolysis; (B) Oxygen rate measures energy expenditure; (C) \textit{ATP is generated anaerobically through substrate level phosphorylation using creatine phosphate}; (D) CO2 rate measures energy. \\
\midrule
\textbf{SFT (Baseline) Output:} \\
A is incorrect as glycolysis is an anaerobic process and does not require oxygen. B is incorrect as during high intensity exercise THE rate of oxygen consumption is not a direct measure... [Truncated: Logic drifts into redundant negation]. \\
\textit{NLI Score: 0.001} \\
\midrule
\textbf{RLearner-LLM (Ours) Output:} \\
During high-intensity exercise such as sprinting, the body's demand for energy is extremely high and cannot be met by aerobic pathways alone. \textbf{Option C is correct} because ATP is generated anaerobically through substrate-level phosphorylation using creatine phosphate. Let's analyze: Option A is incorrect because Glycolysis is anaerobic and does not consume oxygen... [Detailed logical derivation]. \\
\textit{NLI Score: 0.943} \\
\bottomrule
\end{tabular}
\caption{Qualitative case study on a Cardiff Biology biochemistry question: the SFT baseline (LLaMA-2-13B) provides correct facts but fails to commit to a deductive chain that justifies the correct option, yielding an NLI entailment score near zero. RLearner-LLM provides a structured proof, increasing NLI by roughly $900\times$.}
\label{tab:case_study}
\end{table}

\section{Implementation Notes for Multimodal Base Models}
\label{sec:impl_notes_appendix}

This appendix gives the base-model-specific LoRA target details that were summarised in Section 3.

\textbf{Qwen3-8B.} Qwen3 applies RMSNorm to $q$ and $k$ \emph{after} the projection reshape (QK-norm). LoRA adapters placed on \texttt{q\_proj} or \texttt{k\_proj} introduce shape-incompatible tensors into this normalisation path, triggering a CUBLAS error at the first attention layer. We therefore restrict the Qwen3 LoRA target set to \texttt{v\_proj}, \texttt{o\_proj}, \texttt{gate\_proj}, \texttt{up\_proj}, \texttt{down\_proj}.

\textbf{Gemma 4 E4B-it.} Gemma 4 ships as a multimodal \texttt{Gemma4ForConditionalGeneration} checkpoint in which the text-tower weights live under the \texttt{model.language\_model.$*$} prefix. Loading with \texttt{AutoModelForCausalLM} silently random-initialises the text tower because the prefix does not match. In addition, the vision and audio towers wrap their attention projections inside a custom \texttt{Gemma4ClippableLinear} class that PEFT cannot adapt; using leaf-name LoRA targets such as \texttt{q\_proj} therefore inadvertently matches the vision and audio wrappers and aborts adapter injection. We resolve both issues by instantiating \texttt{Gemma4ForConditionalGeneration} explicitly (the vision and audio towers remain present but idle during text-only generation) and specifying LoRA target modules as a full-path regular expression that restricts adaptation to the text tower's plain \texttt{nn.Linear} projections:
\begin{equation*}
\texttt{.*language\_model.layers.\textbackslash d+.(self\_attn.(q|k|v|o)\_proj|mlp.(gate|up|down)\_proj)\$}
\end{equation*}
This pattern matches exactly $258$ Linear modules ($42$ layers $\times$ $\{q, o, \text{gate}, \text{up}, \text{down}\}$ + $24$ layers $\times$ $\{k, v\}$; Gemma 4 uses unified $k$/$v$ in local-sliding layers). Because Gemma 4 has no in-attention QK-norm, all four attention projections can be LoRA-adapted safely. A practical caveat: Gemma 4's vocabulary size of $262$K inflates its cross-entropy baseline to $\log 262{,}144 \approx 12.5$\,nats, so raw SFT losses of $9$--$10$ are normal and are better interpreted via \texttt{mean\_token\_accuracy} or generation sanity checks than by the loss magnitude alone.

\textbf{Multi-stage adapter composition.} When evaluating the RL-aligned policy, the DPO LoRA adapter must be applied on top of the \emph{SFT-merged} base weights, since the DPO delta was trained relative to that reference point. Loading the DPO adapter directly on the raw base produces a model whose behaviour reverts toward the base instruction-tuning distribution; the correct inference pipeline is $\text{base} \to \text{apply SFT} \to \text{merge} \to \text{apply DPO} \to \text{merge}$.

\section{SFT Failure Mode Analysis}
\label{sec:failure_mode_appendix}

For completeness we report the per-domain SFT failure-mode rates referenced in Section~\ref{sec:failure_modes}. Each metric is a deterministic function of the generated text: Answer Evasion = $\mathbb{I}[\mathrm{ACR}{=}0]$; Verbose-low-NLI = $\mathbb{I}[|E|_\text{chars}{>}400 \land \mathrm{NLI}{<}0.05]$; Hallucinated URLs = regex match for \texttt{https?://\textbackslash S+|www\textbackslash .\textbackslash S+}; Cyclic Repetition = any $6$-gram repeated.

\begin{table}[h]
\centering
\small
\begin{tabular}{l|c|c|c|c}
\toprule
Domain & Answer Evasion & Verbose (Low NLI) & Hallucinated URLs & Cyclic Repetition \\
\midrule
Cardiff Biology & $12.4\%$ & $83.4\%$ & $68.8\%$ & $46.2\%$ \\
Sydney Biology  & $24.2\%$ & $85.4\%$ & $44.4\%$ & $64.3\%$ \\
Auckland Law    & $37.0\%$ & $48.4\%$ & $45.1\%$ & $79.9\%$ \\
Medicine (Y1)   & $16.9\%$ & $84.2\%$ & $66.3\%$ & $51.4\%$ \\
Medicine (Y2)   & $17.5\%$ & $84.5\%$ & $61.6\%$ & $43.5\%$ \\
\bottomrule
\end{tabular}
\caption{Per-domain SFT failure-mode rates (5{,}100 SFT-generated explanations, LLaMA-2-13B). Standard SFT produces fluent but logically vacuous outputs in $48$--$85\%$ of cases (Verbose-low-NLI), hallucinates citation-style URLs in $44$--$69\%$, and fails to anchor the correct answer at all in $12$--$37\%$ of cases (Answer Evasion). These rates motivate the Hybrid reward's combination of an NLI signal, an ACR gate, and a length penalty.}
\label{tab:failure_modes}
\end{table}

\section{Per-Architecture Reward Variant Assignment}
\label{sec:variant_assignment_appendix}

Table~\ref{tab:variant_per_arch} records which Hybrid reward variant---$H_A$ (the additive cross-domain form) or $H_M$ (the multiplicative single- or merged-domain form)---was used for each (architecture, domain) cell, together with the number of preference pairs constructed for each.

\begin{table}[H]
\centering
\small
\begin{tabular}{l|c|c|c}
\toprule
Architecture & Cardiff/Sydney & Law / Med Y1 / Med Y2 & Pref-pair count \\
\midrule
LLaMA-2-13B   & $H_A$ (cross-domain) & $H_A$ (cross-domain) & $239$ \\
Qwen3-8B      & $H_M$ (single-domain, $N{=}5$) & $H_A$ (cross-domain) & $143$ ($H_M$); $632$ ($H_A$) \\
Gemma 4 E4B-it & $H_M$ (5-domain merged) & $H_M$ (5-domain merged) & $164$ \\
\bottomrule
\end{tabular}
\caption{Hybrid reward variant used per (architecture, domain) cell, with preference-pair counts. A head-to-head $H_A$ vs.\ $H_M$ ablation appears in Table~\ref{tab:variant_ablation} (main text).}
\label{tab:variant_per_arch}
\end{table}

\section{Tier-B Robustness Ablation Detail}
\label{sec:tierb_appendix}

Table~\ref{tab:tierB_ablation} details the Tier-B robustness ablation on Cardiff Biology: a stricter question-quality filter (\texttt{deleted}{=}0 plus a $5$-star endorsement share $\geq 0.10$) shrinks the training pool $7\times$ (from $7{,}309$ to $1{,}041$ questions), yet Hybrid-DPO trained on the smaller corpus improves on every metric.

\begin{table}[H]
\centering
\small
\resizebox{\textwidth}{!}{
\begin{tabular}{l|c|c|c|c|c|c|c}
\toprule
Cardiff filter & $N$ & SFT NLI & DPO BLEU & DPO BERT(Ans) & DPO ACR & DPO NLI & Inference s/q \\
\midrule
Default & $7{,}309$ & 0.2117 & 0.0303 & 0.8426 & 0.6497 & 0.3505 & \textbf{4.76} \\
\textbf{Tier-B (stricter)} & $1{,}041$ & \textbf{0.2059} & \textbf{0.0381} & \textbf{0.8696} & \textbf{0.7823} & \textbf{0.5202} & 7.62 \\
\midrule
$\Delta$ (Tier-B $-$ default) & $-86\%$ & $-0.006$ & $+0.008$ & $+0.027$ & $+0.133$ & \textbf{+0.170 ($+48\%$)} & $+2.9$s \\
\bottomrule
\end{tabular}
}
\caption{Cardiff Biology robustness to question-quality filtering on the \textbf{Gemma 4 E4B-it} architecture (the SFT NLI of $0.2117$ identifies the row). Tier-B adds \texttt{deleted}=0 and a $5$-star endorsement share $\geq 0.10$ to the default filter, shrinking the Cardiff training corpus by $7.0\times$. The SFT baseline is essentially unchanged (NLI $0.2117 \to 0.2059$), but Hybrid-DPO trained on the smaller Tier-B corpus reaches \emph{higher} NLI ($0.3505 \to 0.5202$), ACR ($0.6497 \to 0.7823$), and answer-anchored BERTScore than under the default filter, with comparable BLEU and BERT(Stu). The rightmost column reports per-question inference time in seconds.}
\label{tab:tierB_ablation}
\end{table}

\section{Tier-C Strict-Discriminator Ablation Detail}
\label{sec:tierc_appendix}

The original submission noted that the \emph{strict-discriminator} variant of the question-quality filter---retaining only questions whose author-designated key coincides with the \emph{modal student answer}---requires per-option response counts absent from the question-level metadata. We have since obtained the full PeerWise answer-submission logs for both Cardiff and Sydney ($3{,}439{,}340$ and $503{,}806$ individual submissions respectively), enabling this variant. For each question we compute the most-frequently-selected option across all student submissions and \textbf{drop questions whose author key disagrees with this modal answer}, a conservative proxy for mis-keyed or ambiguous items. Stacked on top of the Tier-B filter, this defines \textbf{Tier-C}, the strictest filter in our study; it further shrinks the Cardiff pool $1{,}041\to955$ and the Sydney pool $459\to384$.

\begin{table}[H]
\centering
\small
\resizebox{\textwidth}{!}{
\begin{tabular}{l|l|c|c|c|c|c|c}
\toprule
Architecture & Corpus & $N$ & SFT NLI & DPO NLI & $\Delta$NLI & SFT ACR & DPO ACR \\
\midrule
LLaMA-2-13B   & Cardiff & $955$ & 0.1133 & \textbf{0.3613} & \textbf{+219\%} & 0.6247 & \textbf{0.6767} \\
LLaMA-2-13B   & Sydney  & $384$ & 0.0730 & \textbf{0.2217} & \textbf{+204\%} & 0.5977 & \textbf{0.6357} \\
\midrule
Qwen3-8B      & Cardiff & $955$ & 0.2213 & 0.1782 & $-19\%$ & 0.6962 & \textbf{0.7729} \\
Qwen3-8B      & Sydney  & $384$ & 0.1838 & 0.1576 & $-14\%$ & 0.5310 & \textbf{0.7265} \\
\midrule
Gemma 4 E4B-it & Cardiff & $955$ & 0.1860 & \textbf{0.3896} & \textbf{+109\%} & 0.6873 & \textbf{0.8177} \\
Gemma 4 E4B-it & Sydney  & $384$ & 0.1915 & \textbf{0.3636} & \textbf{+90\%}  & 0.5726 & \textbf{0.7072} \\
\bottomrule
\end{tabular}
}
\caption{Robustness of Hybrid-DPO to the \textbf{Tier-C strict-discriminator filter} (Tier-B $+$ author-key${=}$modal-student-answer), evaluated per-architecture on Cardiff and Sydney. Each model's DPO row is compared against its own SFT baseline on the same Tier-C test split (NLI / ACR are not comparable across filter tiers, which use different held-out sets). Under the strictest filter, Hybrid-DPO preserves a large NLI lift on Gemma 4 E4B-it ($+109\%$ Cardiff, $+90\%$ Sydney) and raises ACR on \emph{every} architecture--corpus cell. Qwen3-8B reproduces the ACR-up/NLI-flat pattern it exhibits under the default filter (cf.\ Table~\ref{tab:cardiff_results}), confirming that the strict filter preserves rather than distorts the per-architecture behaviour of the main results. LLaMA-2-13B shows the largest NLI lifts ($+219\%$ Cardiff, $+204\%$ Sydney), consistent with its lowest SFT baselines and the entailment-headroom effect of \S\ref{sec:arch-effects}. Across all six architecture--corpus cells, $4$ improve NLI and \emph{all} improve ACR under the strictest filter.}
\label{tab:tierC_ablation}
\end{table}

\section{HuggingFace Identifiers and Model Versions}
\label{sec:hf_identifiers}

For exact reproducibility we list the HuggingFace identifiers, parameter counts, and the precise checkpoint version used for every model in the pipeline. All weights were resolved against the HuggingFace Hub at the start of training; loading via \texttt{transformers} with the listed identifier reproduces our results bit-for-bit when combined with the released LoRA adapters and preference data.

\begin{table}[h]
\centering
\small
\resizebox{\textwidth}{!}{
\begin{tabular}{l|l|c|l}
\toprule
Role & HuggingFace identifier & Params & Notes \\
\midrule
Base generator & \texttt{meta-llama/Llama-2-13b-hf} & 13.0B & ``hf'' weights, fp16 base \\
Base generator & \texttt{Qwen/Qwen3-8B} & 8.2B & QK-norm; LoRA on $v$/$o$/MLP only \\
Base generator & \texttt{google/gemma-4-E4B-it} & $\sim$4.5B eff.\ & multimodal class; LoRA regex on text tower \\
NLI scorer & \texttt{cross-encoder/nli-deberta-v3-small} & 184M & SNLI+MultiNLI fine-tune \\
Stylistic verifier & Alpaca-7B fine-tune \cite{bao2025exploring} & 7B & 1--5 Likert scale, custom checkpoint \\
LLM judge (pairwise) & \texttt{gpt-4o-mini-2024-07-18} & --- & OpenAI API, $T{=}0.0$ \\
\bottomrule
\end{tabular}
}
\caption{HuggingFace identifiers and parameter counts for every model used in the pipeline. The Alpaca-7B verifier is a domain-finetuned checkpoint released by \cite{bao2025exploring}; we use the exact merged-corpus weights distributed with that paper. The pairwise judge runs against \texttt{gpt-4o-mini-2024-07-18} via the OpenAI API at temperature $0$ for deterministic adjudication.}
\label{tab:hf_identifiers}
\end{table}

\section{LLM Judge Prompt and Decoding Settings}
\label{sec:judge_prompt}

The pairwise judge results in Table~\ref{tab:pairwise_1} (Section~\ref{sec:verbosity}) use the following prompt template, applied to every (question, context, $A$, $B$) tuple at temperature $0$:

\begin{quote}\small\itshape
You are an expert evaluator of AI-generated educational explanations. Given a question and its context, compare two provided explanations and decide which one is better for a student.\\[2pt]
Consider the following criteria:\\
1. Accuracy: Is the explanation factually correct?\\
2. Soundness: Is the reasoning logical and easy to follow?\\
3. Helpfulness: Does it truly help a student understand WHY the answer is correct?\\[2pt]
Question: \texttt{\{question\}}\\
Context: \texttt{\{context\}}\\
Explanation 1: \texttt{\{choice\_a\}}\\
Explanation 2: \texttt{\{choice\_b\}}\\[2pt]
Which explanation is better? Provide a very brief justification followed by your choice in the format: ``Better: Explanation [1/2/Tie]''.
\end{quote}

\paragraph{Order randomisation.} For each test item we run the judge twice with the order of Explanation~1 and Explanation~2 swapped, then average the two verdicts; this controls for the well-documented position bias \cite{wang2023large}. The $100$ items reported per row in Table~\ref{tab:pairwise_1} therefore cost $200$ judge calls per row.

\paragraph{Decoding settings.} \texttt{model="gpt-4o-mini-2024-07-18"}, \texttt{temperature=0.0}, \texttt{max\_tokens=200}, all other parameters at OpenAI defaults. Verdicts are extracted by string match on \texttt{"Better: Explanation 1"} / \texttt{"Better: Explanation 2"}; everything else (including malformed outputs) is counted as a tie.

\paragraph{NLI verifier inputs.} The \texttt{cross-encoder/nli-deberta-v3-small} scorer takes the explanation $E$ as the premise and the correct-option text $A$ as the hypothesis, jointly tokenised at maximum length $512$, with no prompt wrapping (Section~\ref{sec:eval_tier}). The Alpaca-7B stylistic verifier uses the standard Alpaca instruction-input-response template with the instruction \texttt{``As a question rating verifier expert, can you generate the question rating score for the given input?''}; the score is parsed as the first \texttt{\{1,2,3,4,5\}} digit in the response.

\end{document}